\documentclass[runningheads]{llncs}
\usepackage{graphicx}

\usepackage{tikz}
\usepackage{comment} 
\usepackage{amsmath,amssymb} 
\usepackage{color}
\usepackage{url}

\begin{document}
\pagestyle{headings}
\mainmatter
\def\ECCVSubNumber{100}  

\title{1st Place Solution to ECCV-TAO-2020: Detect and Represent Any Object for Tracking} 


\titlerunning{DRAO}
%
\author{Fei Du\inst{1} \and
Bo Xu\inst{2} \and
Jiasheng Tang\inst{1} \and
Yuqi Zhang\inst{1} \and
Fan Wang\inst{1} \and
Hao Li\inst{1}
}
\authorrunning{F.Du et al.}
%
\institute{Alibaba Group. \and
Cainiao Inc.\\\
\email{\{dufei.df,jiasheng.tjs,gongyou.zyq,fan.w,lihao.lh\}@alibaba-inc.com songbai.xb@cainiao.com} }
\maketitle

\begin{abstract}
   We extend the classical tracking-by-detection paradigm to this tracking-any-object task. Solid detection results are first extracted from TAO dataset. Some state-of-the-art techniques like \textbf{BA}lanced-\textbf{G}roup \textbf{S}oftmax (\textbf{BAGS}\cite{li2020overcoming}) and DetectoRS\cite{qiao2020detectors} are integrated during detection. Then we learned appearance features to represent any object by training feature learning networks. We ensemble several models for improving detection and feature representation. Simple linking strategies with most similar appearance features and tracklet-level post association module are finally applied to generate final tracking results. Our method is submitted as \textbf{AOA} on the challenge website\footnote{Due to unknown issue when submitted to MOT server, we failed on archiving all three files first. Then by replacing empty json files for train and validition, it seems good. So if anyone who is interested in our train and validation results, please contact us for this purpose. We would like to share those ones.}. Code is available at \url{https://github.com/feiaxyt/Winner_ECCV20_TAO}.
\end{abstract}

\section{Introduction}
Tracking Any Object (TAO)\cite{2020TAO} dataset is the first work to bring more sophisticated categories in an open world to the multiple-objects-tracking (MOT) community. Besides, despite the whole consecutive frames are given, they are only annotated at 1 FPS, which makes some popular MOT frameworks not work. Since there are over 1200 categories in it, detecting and tracking on such a long-tailed dataset is super challenging. In this paper, we will describe our method used for TAO and our thoughts about tracking on such a hard setting.

We follow the classical tracking-by-detection framework to build TAO pipeline. Since better detection leads to powerful tracking results, especially for this long-tailed task, we carefully built our fundamental detection model with Mask-RCNN. More details can be found in \textbf{2.1}.
In the tasks of multiple objects tracking on pedestrians or vehicles (like KITTI, MOT-16, MOT-17, etc.), tracking is usually done by combining appearance and motion features since it also tracks on real-time videos. But as for TAO, we tested some mainstream tracking approaches within real-time and low-frame-rate ones. We chose to track with anything-appearance feature on 1 FPS. More details can be found in \textbf{2.2} and \textbf{2.3}.

\section{Method Details}
\subsection{Detection}
We trained two detection models with MMDetection\cite{mmdetection} framework from Open-MMLab. The first model is built with a strong-enough backbone network, SENet-154 with RFP, and then fine-tuned with balanced group softmax loss on LVIS\cite{gupta2019lvis} v1.0 dataset. A v1.0-to-v0.5 LVIS label mapping is applied with \textbf{synset} field to make it possible to inference on TAO. Besides, we directly bring an open-source model \textbf{BAGS} to TAO, which is the state-of-the-art method for modeling long-tailed data.

\subsection{Tracking: Category-Agnostic Feature Representation}
Before TAO was released, multiple-objects-tracking has always been performed in a full-frame-rate setting, which means all the frames are used for detection/tracking, no matter in online or offline pipelines. However, most of these previous methods failed on TAO under our tests by using the metric of mAP. Better mAP leads us to pay more attention to the Recall of detections and thus brings some challenges while tracking:
\begin{enumerate}
\item With more than 400 categories, the object motion is highly irregular. Combining with the camera motion, object trajectories shown in those videos are very unpredictable. Even tracking with detections at 30 FPS, SORT-like tracker still cannot get satisfying results. 
\item Compared to MOTChallenge\cite{MOT16}, more boxes from detection methods are desired to increase the recall rate. Therefore, algorithms about linking detection boxes in an offline manner didn't demonstrate strong advantages here, such as Min-Cost Network Flow or Graph Neural Networks. We also trained a GBDT model to use geometry cues and appearance cues for linkage prediction, which only got less than 90\% precision.
\item Compared to Single-Object-Tracking which is usually category-agnostic, there is no user-initialized box. Tracking on such a number of boxes is also costly with high frame rate. 
\end{enumerate}

Thus, the most feasible way is to link detection boxes only based on their appearance features. 
The model used for feature representation is trained with a State-of-the-Art Person-ReID framework, whose task is to find a feature embedding for instance matching of any object category. With collections from a few SOT datasets including YouTube-BB\cite{youtubebb}, GOT-10k\cite{got10k} and ImageNet VID\cite{ilsvrc}, over 200,000 trajectories are obtained, which can be treated as 200,000 instance IDs as the training data. We believe this could be able to cover a large variety of object appearance variation, like lighting conditions, view points, deformations, etc., to help learning a discriminative feature representation for any category of objects. 

Therefore, we train two ReID models to extract features. One is trained using only SOT datasets, and the other is trained by adding the training set of TAO.

Another line of feature representation used in tracking is by training a Siamese-like network inspired by popular Single-Object-Tracking frameworks. Similar to SiamRPN\cite{Li2019SiamRPN}, the model is built to discriminate from similar ones than any other detractors. After feature extraction with backbone net, feature map at the center point with 256-dimensional is selected to represent the given object. During our experiments, this kind of feature doesn't perform as well compared with features from ReID model.

Hence, with the two appearance models, we can link boxes through time. We also found that detecting more boxes on more frames might accumulate errors over time. We thus set our tracking method with 1 FPS.

\subsection{Post-processing and Model Ensemble}

Prior to TAO, \cite{tangmin} shows a direct improvement on MOT dataset by using trajectory-fix, the \textbf{PA} module in their work. We use a similar idea and re-implement it in a simple way: re-linking these two tracklets which have no time-overlap with these tracklet-level mean appearance features. This implementation can reach similar performance but with less cost and easier implementation.

\begin{figure*}[h]
\begin{center}
   \includegraphics[width=0.9\linewidth]{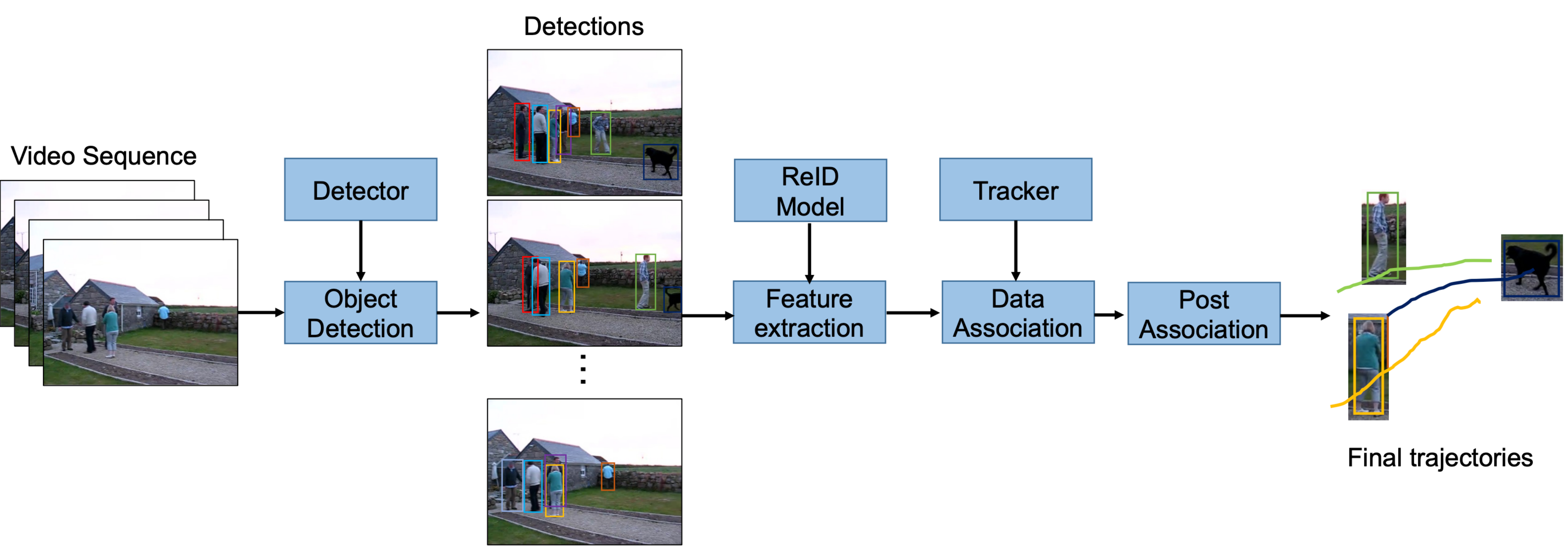}
\end{center}
   \caption{Pipeline of Our Approach}
\label{TAO}
\end{figure*}

As for model ensemble of tracking, we found that directly mixing our tracking results from two detection models can improve the performance of mAP. This can be explained as improving the recall of tracklets. The two appearance features are concatenated during the tracking stage. Recalling the pipeline: the two detection models first detect object respectively. Then appearance features concatenated from two ReID models is applied to track with these two detection results at 1 FPS. Finally, two tracking results are merged together, then the \textbf{PA} module for tracklet-level fix is used for post-processing. Figure \ref{TAO} shows the whole pipeline of our approach.

\section{Experiments}
\subsection{Implementation}
As mentioned above, the detection module is implemented with MMDetection framework. And the appearance model is trained with \cite{luo2019strong}\cite{luo2019bag}. The detailed feature linking strategies come from \textbf{DeepSORT}\cite{Wojke2018deep} by removing its dependency on geometry cues.

The biggest model in our submission is the LVIS v1.0 detection model, which is distributedly trained  with three servers, each with 8 of 32G V100 GPU. It is built with DetectoRS with SENet-154 as backbone, then fine-tuned with (\textbf{BAGS}) for modeling long-tailed-distribution data like LVIS.

\subsection{Results}
Due to limited submission tries on TAO test set, we reported the performance of our detection models on validation set of LVIS and tracking results on validation set of TAO respectively. 

We use \textbf{Model 1} (the open-source BAGS) for our initial experiment. This model can archive \textbf{34.98} of AP50 on TAO validation set, compared with \textbf{27.26} of the \textbf{baseline} model. Composed with \textbf{SORT} at 30 FPS, slight improvement of mAP can be reached. Then we test this model with our anything-appearance features. A discriminative result can be easily reached with our SOT features and ReID ones. Better result can be gained when switched to detection \textbf{Model 2}.

An important trick for detection output before tracking is that only output the 482 LVIS categories appeared in TAO, which would improve the recall of labels considered in TAO during detection phase. We mark it as \textbf{Model-\{number\}-482} in the table. Table \ref{detection} and \ref{tracking} show the performance in detail.
\textbf{SOT} and \textbf{ReID} are short for SOT appearance feature and anything-appearance feature respectively.

We also analyse our models with oracle setting for knowing the upper bound of our approach. The results are illustrated in table \ref{oracle}. As can be seen from the table, we achieve performance of 46.10 mAP and 48.51 mAP by linking the detection results of \textbf{Model 1} and \textbf{Model 2} with oracle tracks, respectively. However, by combining the detection results of the two models, we achieve much higher performance of 57.72 mAP, which sets a high upper bound for our approach.

\begin{table*}
\begin{center}

\caption{Detection performance on LVIS v0.5 validation set}
\label{detection}
\begin{tabular}{l   l   l   l}
\hline\noalign{\smallskip}
\textbf{Model Name} & \textbf{Main Setting} & \textbf{Dataset for training} & \textbf{mAP} \\
\noalign{\smallskip}
\hline
\noalign{\smallskip}
Model-1 & gs-htc-dconv-x101-64x4d-fpn & LVIS v0.5 {\&\&} COCO & 37.71\\
Model-2 & gs-htc-dconv-senet154-rfp & LVIS v1.0 & 39.70\footnotemark[1] \\
\hline
\end{tabular}
\end{center}
\end{table*}
\footnotetext[1]{Reported on LVIS v1.0 validation set.}

\begin{table}
\tabcolsep=3pt
\begin{center}
\caption{Tracking performance on TAO validation set with various setting}
\label{tracking}
\begin{tabular}{l   l   l   l }
\hline\noalign{\smallskip}
\textbf{Detection} & \textbf{Tracking} & \textbf{Spec.} & \textbf{mAP} \\
\hline
Model-1 & SORT & 30 FPS & 14.80 \\
Model-1 & SOT  & 1 FPS & 23.42 \\
Model-1 & ReID1   & 1 FPS & 24.52 \\
Model-1-482 & ReID1  & 1 FPS & 25.02 \\
Model-1-482 & ReID2  & 1 FPS & 25.53 \\
Model-1-482 & ReID1+ReID2  & 1 FPS & 26.02 \\
Model-2-482 & ReID1+ReID2 & 1 FPS & 26.32 \\
Model-1-2-482 & ReID1+ReID2 & 1 FPS & 28.59 \\
Model-1-2-482 & ReID1+ReID2+PA & 1 FPS & 29.27 \\
\noalign{\smallskip}
\hline
\end{tabular}
\end{center}
\end{table}

\begin{table}
\tabcolsep=20pt
\begin{center}
\caption{Tracking performance of applying the `track' oracle on TAO validation set.}
\label{oracle}
\begin{tabular}{l   l   l   l }
\hline\noalign{\smallskip}
\textbf{Detection}  & \textbf{mAP} \\
\hline
Model-1-482  & 46.10 \\
Model-2-482 & 48.51 \\
Model-1-2-482 & 57.72 \\
\noalign{\smallskip}
\hline
\end{tabular}
\end{center}
\end{table}


\section{Conclusions}

In our solution for TAO competition, tracking is performed simply with a relatively strong appearance features. The evaluation metric \textbf{mAP} is crucial to this part. Most of existing MOT framework fail on this situation of high recall. The only paradigm that we didn't test and believe work with high frame rate is \textbf{tracktor}\cite{tracktor_2019_ICCV}. With huge amount of data and limited time, we cannot test it since the fundamental detection model would be too large and costly to track with real-time videos.

%
%

\bibliographystyle{splncs04}
\bibliography{tao}
\end{document}